\documentclass[10pt,twocolumn,letterpaper]{article}

\usepackage{cvpr}
\usepackage{times}
\usepackage{epsfig}
\usepackage{graphicx}
\usepackage{amsmath}
\usepackage{bm}
\usepackage{amssymb}
\numberwithin{equation}{section}
\numberwithin{figure}{section}
\usepackage{algorithm}
\usepackage[noend]{algpseudocode}
\usepackage{algorithm}
\usepackage[noend]{algpseudocode}
\makeatletter
\def\BState{\State\hskip-\ALG@thistlm}
\makeatother

\newcommand*\samethanks[1][\value{footnote}]{\footnotemark[#1]}

\usepackage[pagebackref=true,breaklinks=true,letterpaper=true,colorlinks,bookmarks=false]{hyperref}

\cvprfinalcopy 

\def\cvprPaperID{****} 

\ifcvprfinal\fi
\begin{document}

\title{Joint Mapping and Calibration via Differentiable Sensor Fusion}

\author{Jonathan P. Chen\thanks{Equal contribution}\\
Uber AI Labs\\
\and
Fritz Obermeyer\samethanks\\
Uber AI Labs\\
\and
Vladimir Lyapunov\\
Uber ATG\\
\and
Lionel Gueguen\\
Uber ATG\\
\and
Noah D. Goodman\\
Stanford University\\
Uber AI Labs\\
{\tt\small \{jpchen, fritzo, vl, lgueguen, ndg\}@uber.com}
}

\maketitle

\begin{abstract}
We leverage automatic differentiation (AD) and probabilistic programming languages to develop an
end-to-end optimization algorithm for batch triangulation of a large number of unknown objects.
Given noisy detections extracted from noisily geo-located street level imagery without depth
information, we jointly estimate the number and location of objects of different types,
together with parameters for sensor noise characteristics and prior distribution of objects
conditioned on side information.
The entire algorithm is framed as nested stochastic variational inference.
An inner loop solves a soft data association problem via loopy belief propagation; a middle
loop performs soft EM clustering using a regularized Newton solver (leveraging an AD framework);
an outer loop backpropagates through the inner loops to train global parameters.
We place priors over sensor parameters for different traffic object types, and demonstrate
improvements with richer priors incorporating knowledge of the environment.

We test our algorithm on detections of road signs observed by 
cars with mounted cameras, though in practice this technique can be used for
any geo-tagged images.
We assume images do not have depth information (\eg from lidar or stereo cameras).
The detections were extracted by neural image detectors and classifiers, and we
independently triangulate each type of sign (\eg ``stop", ``traffic light").
We find that our model is more robust to DNN misclassifications than current methods, 
generalizes across sign types, and can use geometric information to increase
precision (\eg Stop signs seldom occur on highways).
Our algorithm outperforms our current production baseline based on $k$-means clustering.
We show that variational inference training allows generalization by learning
sign-specific parameters.
\end{abstract}

\section{Introduction}

\noindent
\begin{figure}[t]
\begin{center}
	\includegraphics[width=0.9\linewidth]{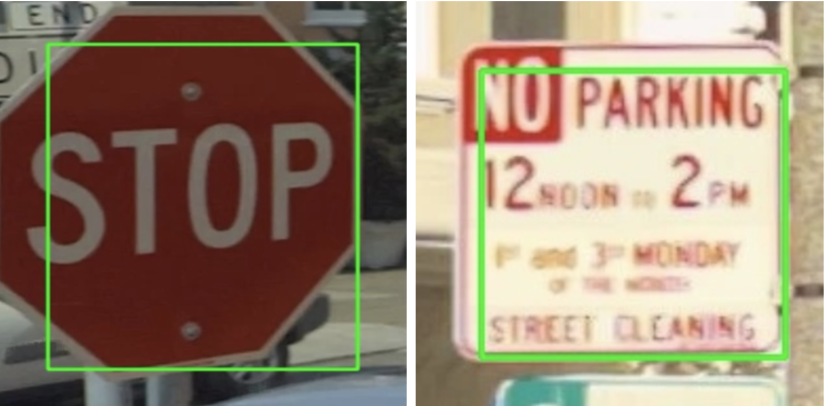}
\end{center}
    \caption{Sample sign detections from photos taken from cameras on vehicles.}
\label{fig:sign}
\label{fig:onecol}
\end{figure}

One of the most challenging problems in building autonomous vehicles and route planning is constructing
accurate maps. These maps are often algorithmically constructed from a combination of
sources of information including satellite imagery, government data, street view imagery,
and human labeling \cite{wegner2016cataloging,mattyus2016hd}.  These sources trade off cost and scalability with accuracy; as with most
problems with big data, human labeling is accurate, but expensive.  Therefore,
improving the accuracy of automatic systems that can perform robust mapping in the presence
of noisy detection and classification of objects would save both cost and effort.

%

$K$-means clustering is a common algorithm used for clustering data that come from different
sources ~\cite{yu2012optimized,gonen2014localized}. In the case of sensor fusion, the algorithm partitions
data by assigning nearby objects to the nearest cluster using a metric such as Euclidean distance.  There 
has been work in developing heuristic algorithms (\eg Lloyd's Algorithm \cite{lloyd1982least}) for
improving clustering in certain
domains.  For the problem of sensor fusion, it is often overconfident about false detections,
as it fails to incorporate prior information and uncertainty about observation parameters
such as observer location.

We propose a probabilistic model that performs bundle adjustment with road signs at scale, learning
measurement parameters, \eg observable radius and GPS error, while triangulating the 
probable locations of signs through an agglomerative clustering algorithm.  This algorithm
performs expectation maximization (EM) via message passing and Newton's method.  Critically,
the clustering solver is differentiable, so its parameters can be learned jointly with
stochastic variational inference (SVI).  Optimization
by inference requires a generative model, \ie a model that describes how latent variables
produce observed data. The generative model generates clusters given ray origins and directions
given camera and neural net confidence parameters.  It is used in conjunction with the clustering
solver in a variational inference setting to optimize the global parameters. 

We show that precision and recall can be increased with domain-specific priors
and by incorporating data such as road networks.
Since not every geographic region has both road network data and labeled ground truth, we learn
{\it maximum a posteriori} (MAP) estimates of geographic parameters offline with limited data,
which can be used to generalize to signs in different geographic regions.

Our contributions are as follows:
\begin{itemize}
	\item We implement a differentiable soft clustering algorithm that uses
      loopy belief propagation (BP) to solve a data association problem
		  and a differentiable Newton solver to predict cluster locations.
	\item We propose a generative model and an approximate inference model
		  that is used to learn model parameters end-to-end on partially-labeled data.
	\item We learn parameters unique to each type of traffic object, and incorporate
		  prior information in the form of road networks, learning \eg
		  where each type of traffic object is typically located w.r.t roads.
	\item We develop heuristics such as sparsification, eccentricity pruning, and locally sensitive
		  hashing to scale to data batches of over 10,000 rays and 1000 objects on each compute node.
	\item We evaluate our method against SOTA methods used in production to build maps at scale.
\end{itemize}

\section{Related work}

Our system jointly performs object clustering, 3-D triangulation
\cite{hartley1997triangulation}, and Bayesian training of parameters for
sensor noise characteristics and object distributions.

Joint tracking and calibration has a long history, \eg  Lin et al ~\cite{lin2004exact}.
Using bearings-only sensors, Houssineau et al. \cite{houssineau2016unified} developed a
unified approach to multi-object triangulation and parameter learning for sensor calibration.
Ristic et al. \cite{ristic2013calibration} developed a joint algorithm for multi-target
tracking and sensor bias estimation based on the Probability Hypothesis Density (PHD) filter.

The PHD filter \cite{mahler2007phd} and labelled multi-Bernoulli filter \cite{reuter2014labeled}
represent collections of unknown number of objects with unknown positions; our representation
can be seen as a Laplace approximation to a multi-Bernoulli filter. To track multiple objects with
unlabeled detections, Williams and Lau \cite{williams2014approximate} solve the data association
problem by using loopy belief propagation \cite{murphy1999loopy} to produce an approximate soft
assignment of detections to objects. Turner et al. \cite{turner2014complete} describe a similar
loopy BP-based multi-object tracking system, formulating the tracker as a variational inference
method \cite{blei2017variational}, similar to our formulation.

The recent availability of automatic differentiation frameworks like PyTorch \cite{paszke2017automatic}
has led to more end-to-end learning approaches in localization \cite{karkus2018particle} and tracking
\cite{jonschkowski2018differentiable}. One crucial advance has been the ability to differentiate through
solutions of optimization problems \cite{gould2016differentiating,amos2017optnet} to enable nested optimization.

\subsection{Variational inference}

Probabilistic inference intends to infer distributions or values of latent
variables $z$ given observed variables $x$ according to a probability
distribution $p(z|x)=p(x,z)/p(x)$.
Variational inference \cite{blei2017variational} is an approximate inference
technique that treats probabilistic inference as an optimization problem by
fitting an approximate distribution $q(z|x;\theta)$ to the model $p(x,z)$ by
maximizing the evidence lower bound (ELBO)
\begin{gather}
    \mathbf{ELBO} = \arg\max_\theta \mathbb E_{\bm{q}(z)} \bigl[ \bm{p}(x,z) - \bm{q}(z|x;\theta) \bigr]
\end{gather}
When variational parameters are shared across data, variational inference is
amenable to stochastic optimization via minibatching (stochastic gradient variational
Bayes \cite{kingma2013auto}) and random sampling of latent variables (stochastic 
variational inference \cite{hoffman2013stochastic}).

Selecting an appropriate variational distribution is an open research problem 
and is often subject to the particulars of a given problem.
A common variational family is the {\it mean-field} variational family, which
imposes independence among the latent variables, \ie the variational distribution factorizes completely:
\begin{gather}
  q_{\text{mean\_field}}(z) = \prod_i q(z_i)
\end{gather}

Variational inference has recently become easier to scale to complex
models through the use of automatic differentiation frameworks \cite{paszke2017automatic}
and high level probabilistic programming languages \cite{carpenter2017stan,bingham2018pyro}.
These tools can compute multiple derivatives through complex control flow, and leverage a number of
techniques for variational inference including the reparameterization trick, Rao-Blackwellization,
and automatic collapsing of discrete latent variables.
For example, whereas previous optimization approaches have leveraged robust least squares solvers
\cite{lourakis2004design}, AD frameworks permit regularized Newton solvers on symbolically computed
likelihood functions, hence permitting a wider class of likelihoods and loss functions in models.

Probabilistic programming languages (PPLs) \cite{carpenter2017stan,bingham2018pyro} generalize
probabilistic graphical models (PGMs) by allowing control flow, recursion, and other high level programming features in probabilistic models.
A probabilistic program with static single assignments and no control flow corresponds to a
probabilistic graphical model.

\section{Method}
\subsection{Joint probabilistic data association}
The data assignment of rays to clusters is solved with an EM algorithm, iteratively alternating
between computing expectations of assignments and maximizing object locations at each step.
Our EM algorithm consists of two phases: during the {\it E}-step, we use loopy
belief propagation to compute the marginal association probabilities of assignments
from detections (rays) to objects.

For each object $i \in \{1, ..., n\}$ let $e_i \in \{0, 1\}$ be the existence
variable
which is $1$ if the detection exists and $0$ otherwise.  Similarly, let $a_{ij} \in \{0, 1\}$ be the 
assignment variable which is $1$ if measurement $i$ is assigned to detection $j$ and $0$
otherwise.  False detections are incorporated in $a_{ij}$ by assigning measurements to
a ghost cluster if they associate with no detections. We can think about this setup as a 
bipartite graph with detections and clusters as nodes and assignments as edges.  Then 
the joint probability of the existence and assignment logits is as follows:
\begin{gather}
P(e,a) \propto \gamma (e,a) \prod_{i}\psi_{i}^{e_i}\prod_{ij} \psi_{ij}^{a_{ij}}
\end{gather}
where
\begin{gather}
  \psi^{e_i} = \alpha_i \prod_j (1-p_d(x_i, j))\\
  \psi^{a_{ij}} = \delta_{i, j} \frac{f(z_j \mid x_i)}{f_{\rm FD}(z_j)}\\
  \gamma(e, a)=\begin{cases}
    1, & \text{if $a_{i,j} \leq e_{i,j} \forall i,j$}.\\
    0, & \text{otherwise}.
  \end{cases}
\end{gather}
and $\alpha_i$ is the existence logit, $\delta_{i, j}$ is the detection logit,
and $f/f_{\rm FD}$ is the density ratio of the assignment logits.

The pairwise marginals for each edge in the graph can be approximated by loopy BP.
We define $\mu_{ij}$ to be the message passed from $e_i \rightarrow a_{ij}$ and $\nu_{ij}$
to be the message passed in the reverse direction: $a_{ij} \rightarrow e_i$.  The messages 
being passed are as follows:
\begin{gather}
\mu_{ij} = \frac{\psi_i \prod_{k\neq j}\nu_{kj}}{1 + \psi_i \prod_{k\neq j}\nu_{kj}}\\
\nu_{ij} = 1 + \frac{\psi_{ij}}{1+\sum_{l \neq i} \psi_{lj} \mu_{lj}}\\
\overline{e_i} = \frac{\psi_i \prod_j \nu_{ij}}{1 + \psi_i \prod_j \nu_{ij}}\\
\overline{a_{ij}} = \frac{\psi_{ij}\mu_{ij}}{1 + \sum_k \psi_{kj} \mu_{kj}}
\end{gather}

This algorithm converges quickly; in our experiments, we run the loopy BP
algorithm for only 5 iterations.
We then perform the {\it M}-step with a regularized Newton solver and update cluster
locations and merge nearby clusters within a certain radius.
Instead of a least squares solver, we use a twice differentiable log likelihood and
directly apply a regularized Newton step as in equation \eqref{newton}. This also
allows us to add in other log-likelihood terms such as a geographic prior.
\begin{gather}
  \mathbf{x}_{n+1} = \mathbf{x}_n - [\mathbf{H} + \lambda \mathbf{I}]^{-1} \nabla f(\mathbf{x}_n) \quad \forall \ n \ge 0.
    \label{newton}
\end{gather}
\noindent ${\bf H}$ is the Hessian and $\lambda$ is the regularization factor.
The matrix solve is inexpensive because ${\bf H}$ has block diagonal structure with
blocks of size only 2 or 3 (for 2D or 3D mapping, respectively); hence we can run
this jointly over a tensor of thousands of clusters.
The optimum found by the Newton solver is differentiable with respect to the solver
inputs \cite{gould2016differentiating}
so we can backpropagate through the solution to later learn global parameters.

\begin{algorithm}
\caption{EM Clustering}\label{alg:em}
\begin{algorithmic}[1]
\State {\bf input} $R$
\State $x \gets \textrm{init}(R)$
\While{not converged}:
\State {\it // E-step:}
\State $p_a, p_{ae} = \textrm{LoopyBP}(\log p_d(x, R))$
\State {\it // M-step:}
\State $loss = \sum_{i,j}p_{ae}(i, j)\log p_d(x_i, R_{j})$
\State $x_n \gets \textrm{NewtonStep}(loss, x)$
\State $x_n \gets \textrm{merge(prune(}x_n))$
\EndWhile
\end{algorithmic}
\end{algorithm}
\begin{algorithm}

\caption{LoopyBP}\label{alg:bp}
\begin{algorithmic}[1]
\State {\bf input} $p_a$
\State $\mu_f, \mu_b = 0$
\While{not converged}:
\State $\mu_f \gets \log(1-\exp(\mu_f - \rm{sum}(\mu_b) - p_a))$
\State $p_{ae} \gets \exp(\mu_f+p_a)$
\State $\mu_b \gets \log(1+\exp(p_a-\log(1+\exp(\rm{sum}(p_{ae})-p_{ae}))))$
\EndWhile
\State \Return $p_a, p_e$
\end{algorithmic}
\end{algorithm}

\begin{algorithm}
\caption{NewtonStep}\label{alg:newton}
\begin{algorithmic}[1]
\State {\bf input} $x, loss$
\State $g \gets \mathbf{grad}(loss)$
\State $H \gets \mathbf{grad}(g_1 \ldots g_n)$
\State $H_{\rm reg} \gets H + g / r - \lambda_{\rm min}$
\State $x_{\rm new} \gets x - H^{-1}_{\rm reg}g$
\State \Return $x_{\rm new}$
\end{algorithmic}
\end{algorithm}

\noindent  For our experiments, we take one Newton step after loopy BP
has converged and we run Algorithm ~\ref{alg:em} for 10 iterations. Our
implementations are open source \footnote{\url{http://docs.pyro.ai/en/dev/contrib.tracking.html}}.

\subsection{Inference}
\begin{figure}[t]
\begin{center}
	\includegraphics[width=\linewidth]{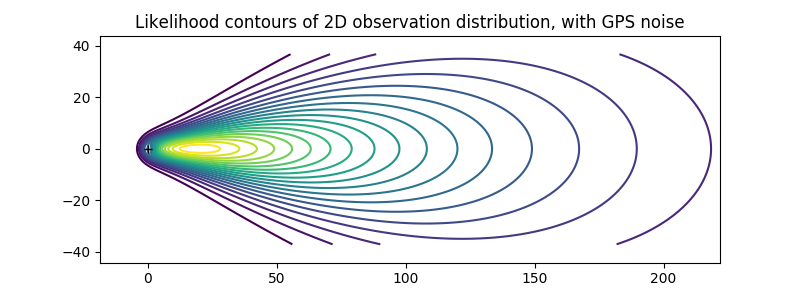}
\end{center}
   \caption{Observable distribution with GPS error and a radius of 50 meters.}
\label{fig:gps}
\label{fig:onecol}
\end{figure}

Our sensor model incorporates two components:
a radial component to model obstruction and invisibility of distant objects (modeled as an Exponential distribution); and
an angular component to model a combination of orientation error of the sensor platform
and segmentation error in the deep object detector.
We do not attempt to estimate the true pose of each camera, since our partitioned
data seldom leads to more than
one detection per camera frame; this is in contrast to traditional bundle adjustment
algorithms that can jointly estimate pose from multiple detections per image.
We instead account for GPS error by approximately convolving our radial-angular
likelihood by a Gaussian.
To make this easier to compute, we preserve the Exponential radial component and
model the angular component as a radius-dependent Von Mises distribution.
The resulting 3-parameter distribution is shown in figure ~\ref{fig:gps}.
During SVI training we learn all three parameters.

To train with variational inference, we need to minimize the Kullback–Leibler (KL) divergence
between the true posterior under our model prior and the variational distribution produced by the
assignment solver.
The ELBO we maximize (equivalent to minimizing the KL) can be written as:

\begin{gather}
    \mathbf{ELBO} = \mathbb{E}_{r, \phi \sim q}[\log \bm{p}_\theta(e, a, x) - \log \bm{q}_\theta(e, a)]
\end{gather}
\noindent where $\bm{q}$ is the distribution produced by the clustering solver parameterized by
$\theta$, and $e$ and $a$ are 
existence and assignment variables respectively.  In this training scheme, SVI is the outermost
optimization algorithm,
taking one gradient step per multiple iterations of the EM algorithm.
Because of the Newton solver's quadratic convergence rate, we can run without gradients
(``detached") in all but the final iteration, and propagate gradients back through only the final
output of the loopy BP. This property is especially helpful because extra clusters at
early iterations are often pruned or merged by the final iteration.

The assignment solver we use to produce the variational distribution generates soft
assignments of the rays to clusters.  These uncertainty estimates are useful when 
making predictions, especially in the context of building maps for autonomous vehicles.
During training we make a mean-field approximation that each object's assignment to rays
is independent of other object's assignments, so that the assignment distribution
factors into independent Categorical distributions.
This approximation allows us to exactly marginalize out assignments, leading to
lower-variance gradient estimates than if we had used Monte Carlo sampling.
This practice is common in soft-assignment EM algorithms.

We approximate the generative process of the clusters as a multi-Bernoulli process \cite{reuter2014labeled}.
We experiment with two prior densities of objects:
first a uniform density over the geographic region, and
second a Spike-and-slab distribution discussed in section \ref{sec:inter}.

Since our data is largely unlabeled, we train in a semisupervised manner.  Specifically,
in areas with labeled ground truth, we probabilistically assign detections to known objects,
assuming all objects are accounted for in ground truth.
In areas without ground truth, we predict candidate clusters via Algorithm ~\ref{alg:em}.
Note that even though the ground truth is known in certain areas, it only provides the candidate
clusters; the data association problem is still unsupervised.

\subsection{Heuristics}

The number of possible assignments grows quadratically with the number of clusters.
To reduce cost, we employ an initialization scheme similar to probabilistic space carving,
whereby we initialize candidate clusters along ray intersections.  By rasterizing the
image and initializing along regions with concentrated rays, we can reduce the 
computation time of the clustering algorithm.

\begin{figure}[t]
\begin{center}
	\includegraphics[width=\linewidth]{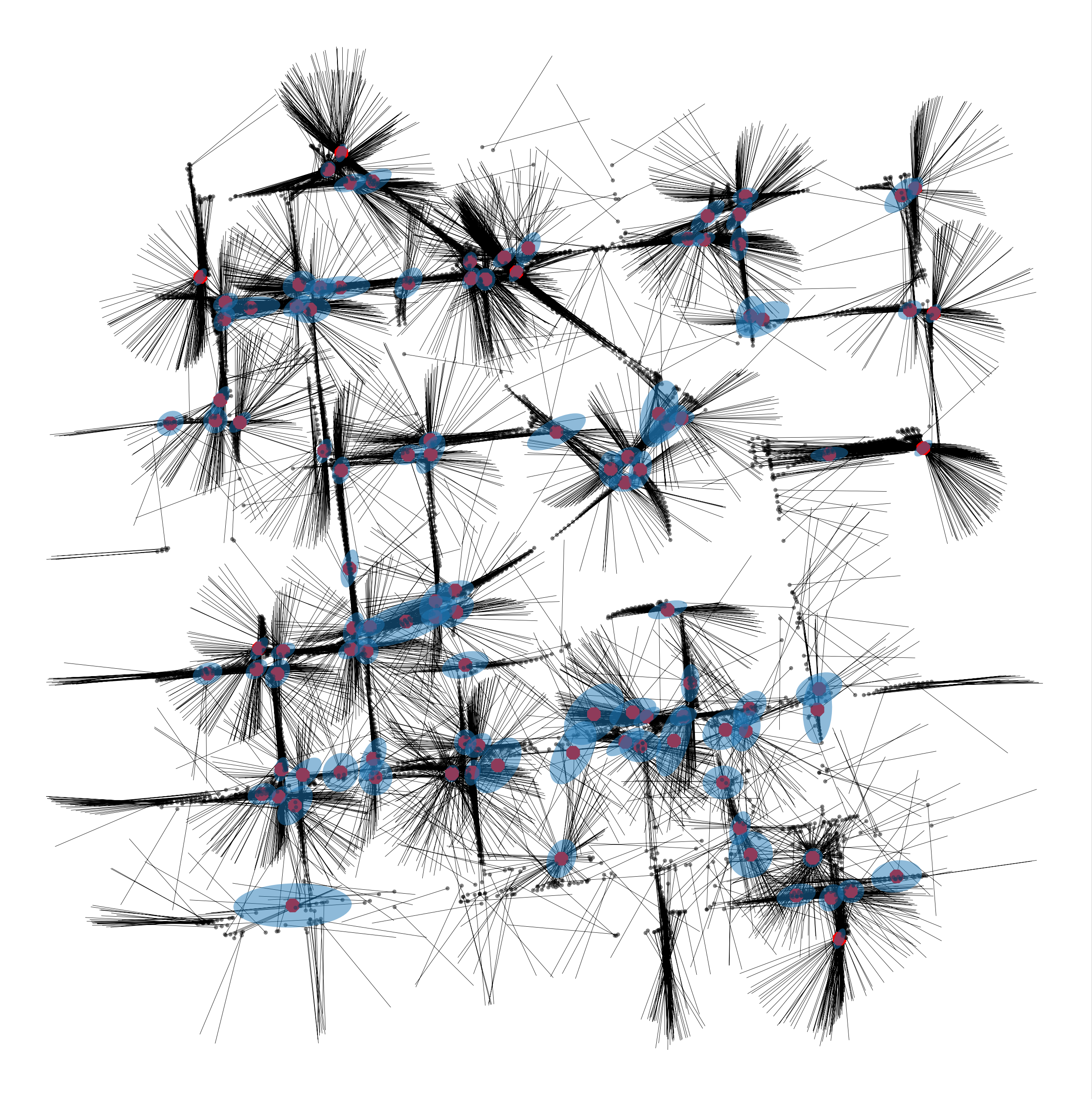}
\end{center}
   \caption{Ellipses of predicted clusters from the eccentricity of assigned rays.}
\label{fig:ellipse}
\end{figure}
To eliminate false detections that arise when two rays ``look past each other"
(\ie lie on the same line in opposite directions),
we compute the eccentricity of the detection from the eigenvalues of its assigned
rays as shown in figure ~\ref{fig:ellipse}.  We then
prune clusters based on an eccentricity threshold. The intuition is that a true
detection (especially one that is not eclipsed by buildings) will be observable from
multiple angles.  Naturally, this varies per sign type since the visibility of signs are subject
to their location and surroundings.  We do something similar with the covariance matrices of the
clusters as well, thresholding them with a variance threshold.

Because the ray-cluster assignment problem is very sparse, we implement sparse tensor versions of
the EM algorithm and BP algorithm, simultaneously assigning tens of thousands of rays to thousands
of clusters on multi-CPU or GPU machines. This is critical for scaling up training on GPUs.
We also implement locally sensitive hashing to efficiently merge nearby clusters greedily.

\section{Experiments}
\begin{figure*}
\begin{minipage}[b]{.45\linewidth}
\includegraphics[width=\linewidth]{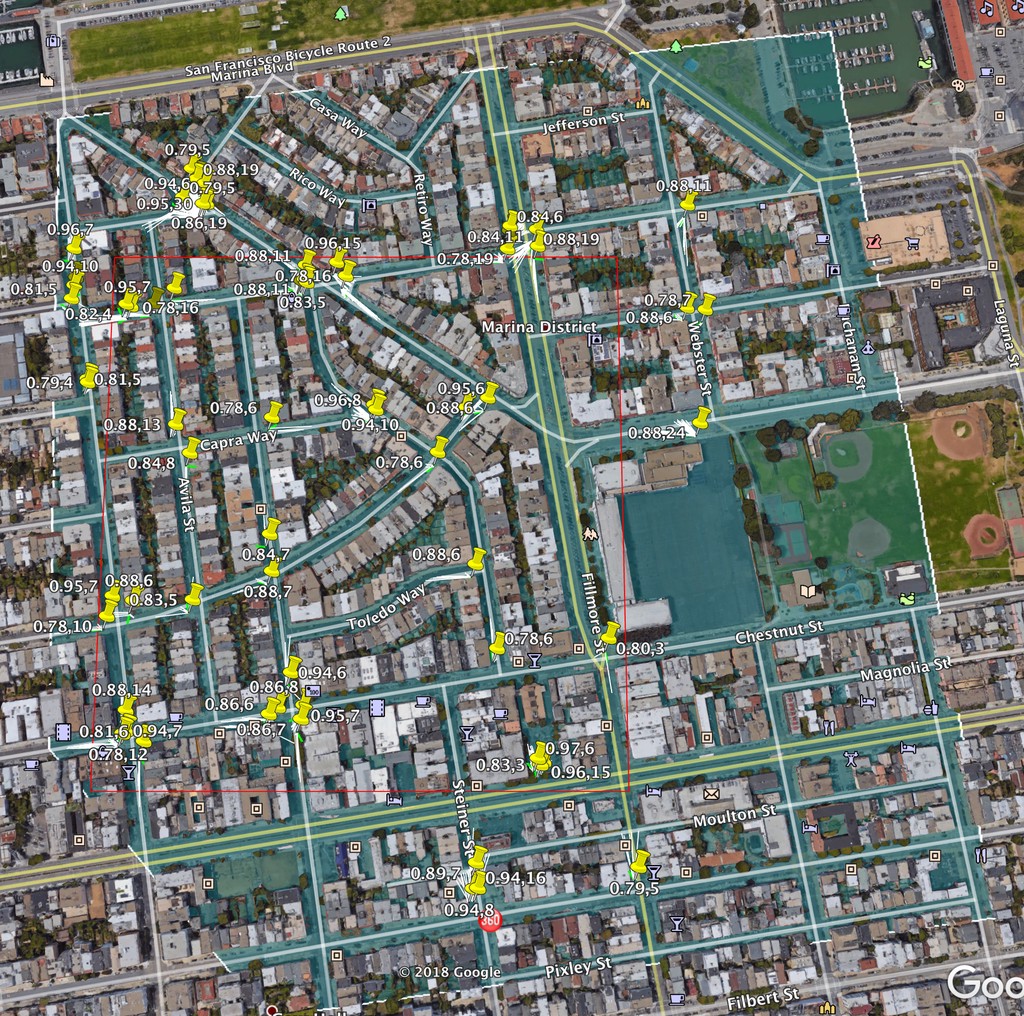}
\end{minipage}\qquad
\begin{minipage}[b]{.45\linewidth}
\includegraphics[width=\linewidth]{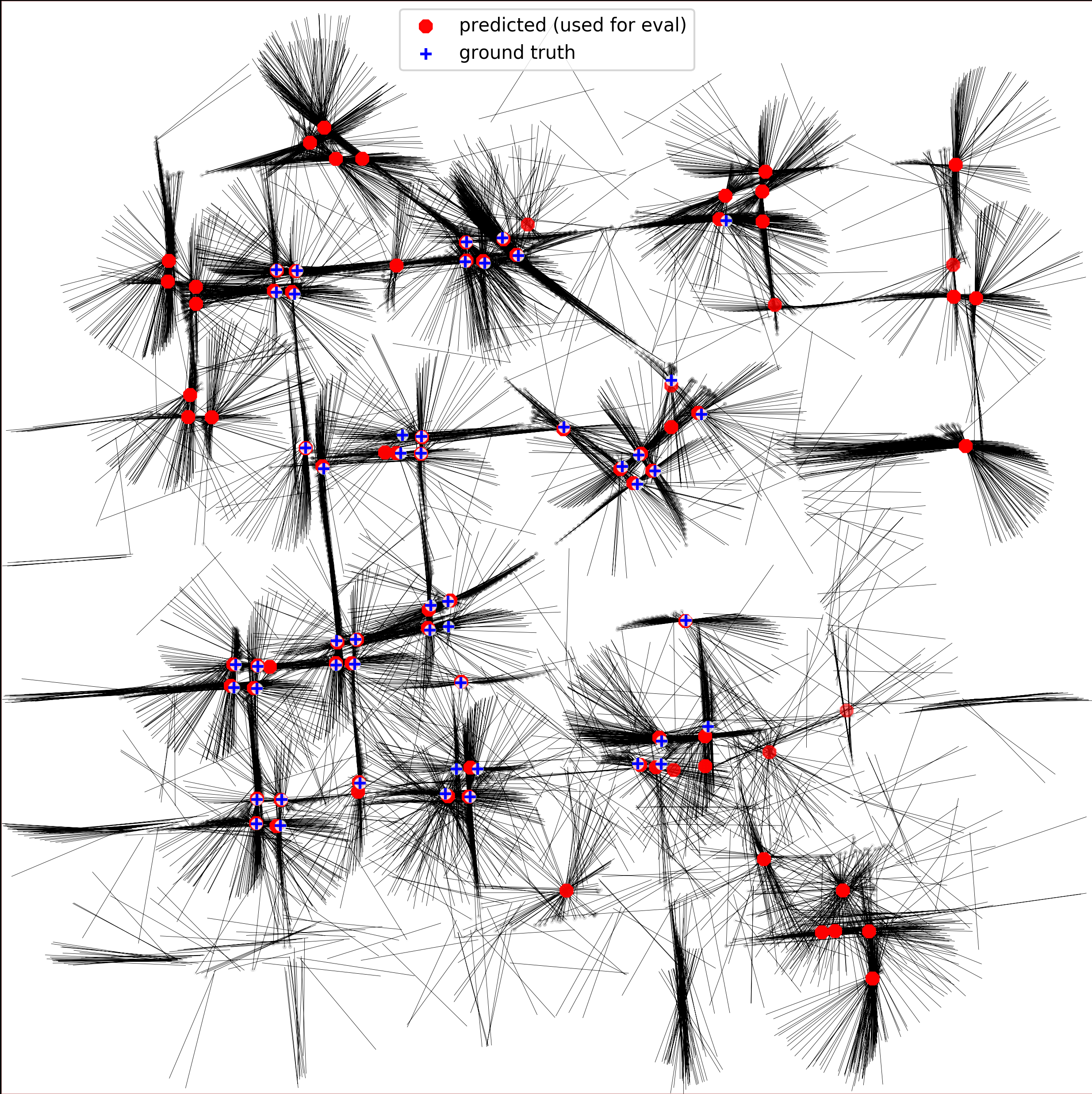}
\end{minipage}\qquad
\begin{center}
\caption{Predicted Stop sign locations (in yellow on the left and red on the right) 
for a 5x5 block region in San Francisco, CA.
Note in the figure on the right that the ground truth labels (blue crosses) for Stop signs 
are incomplete in the region.  The assignment solver has no knowledge of the ground truth clusters.}
\end{center}
\label{fig:blocks}
\end{figure*}

\begin{figure}[ht!]
\begin{center}
	\includegraphics[width=\linewidth]{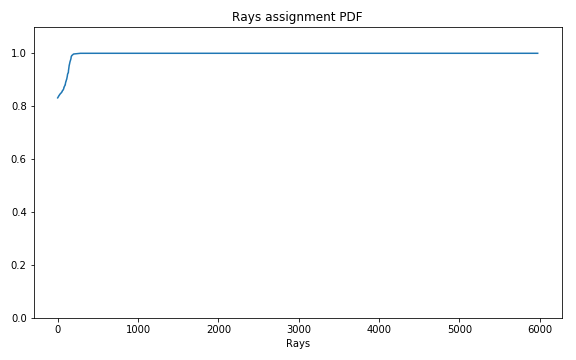}
\end{center}
   \caption{CDF of assignment probabilities of rays for Stop signs.}
\label{fig:cdf}
\label{fig:onecol}
\end{figure}

We run the clustering algorithm with both initial hand-tuned and trained parameters,
and compare precision-recall and F1 metrics against the $K$-means clustering baseline.
We consider each cluster within a 10 meter radius of a ground truth object as a true
positive and additional clusters inside or outside the radius as false positives.
We implement all algorithms in PyTorch \cite{paszke2017automatic} and Pyro \cite{bingham2018pyro}.

\subsection{Data}
The main input data consists of rays denoted by the origin GPS position, the direction of the detection
as a unit vector, and a confidence score from the DNN classifier.  We focus primarily on Stop
signs since they are the canonical case of sign detection and among the most critical for 
autonomous vehicles and ETA estimation.  We look at 
three regions with ground truth in San Francisco: one four-way intersection, a 5x5 block
area, and across a city (approximately ten 5x5 blocks of data). For Stop signs, a typical 5x5 block
contains roughly 6000 rays and may contain anywhere from 60 to 100 Stop signs.  We also 
look at other sign types in these regions with ground truth data since
they have different detection accuracies, characteristics, and/or typical locations than Stop signs.
This will allow us to measure how well SVI generalizes to different sign types.

\subsection{Evaluation}
We evaluate our models against the baseline by looking at three key metrics: precision,
recall, and area under the curve (AUC), which is the integral of the precision-recall
curve and our primary metric of interest.  We look at two regions in particular: the 5x5
block and city level in San Francisco.
The city data consists 10 non-contiguous 5x5 blocks.  We threshold the output clusters below
a certain probability before calculating metrics, using the same threshold across all
three models.

\subsection{SVI training}
We train 8 parameters for every sign type concurrently; since sign-specific parameters are
independent, the training scheme parallelizes easily.  We train only in regions
where we have some ground truth, though even then, ground truth coverage is only partial
in these regions.  Training is subject to the coverage and accuracy of 
the ground truth, since SVI will attempt to explain away rays within a truth region that
are not associated with a true cluster.  In areas where the true clusters are sparse \eg
Crosswalk signs, we notice that SVI training does not make improvements over the manual initialization
scheme.

We incorporate pruning thresholds as non-trainable parameters during training but loosen them
during prediction especially for signs with sparse numbers of rays (\eg Pull through,
Crosswalk signs) since these scenarios lack enough rays for eccentricity and variance to be
a meaningful indicator of the quality of the detection.

Discrete latent variables such as the Categorical variables in the assignment
distribution are known to produce high variance gradient estimates
\cite{tucker2017rebar, grathwohl2017backpropagation}.
To obtain lower-variance gradient estimates,
we enumerate out discrete variables, performing exact inference for discrete latents 
in both our model and the variational approximation.  This eliminates gradient estimator variance due to sampling latent variables, so that the only remaining source of variance during training is the random subsampling of data minibatches.

We train with the Adam optimizer \cite{kingma2014adam} using a learning rate of 0.001
and anneal the learning rate with a decay factor of 0.7.
We partition data into minibatches of approximately 5x5-block regions that contain anywhere
from 40 to 7000 rays each.
We run 5 iterations of loopy belief propagation and 10 EM iterations per SVI step.

SVI learns location-varying parameters such as GPS variance, which may be subject to
the location of the vehicle.  We assume GPS variance to be static and possibly slowly varying in space \eg 
vary between residential and industrial areas of cities due to varying effects of occlusions due to buildings.

\subsection{Quantitative results}
\begin{figure}[t]
\begin{center}
	\includegraphics[width=0.9\linewidth]{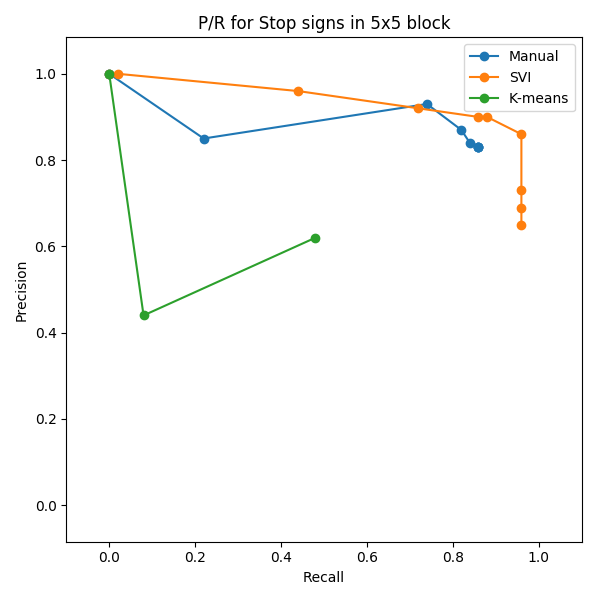}
\end{center}
    \caption{Precision-recall plot for Stop signs in a 5x5 region. Note the improvement in AUC for the SVI-tuned model.}
\label{fig:pullthru}
\label{fig:onecol}
\end{figure}

\begin{figure*}
\begin{minipage}[b]{.45\linewidth}
\includegraphics[width=\linewidth]{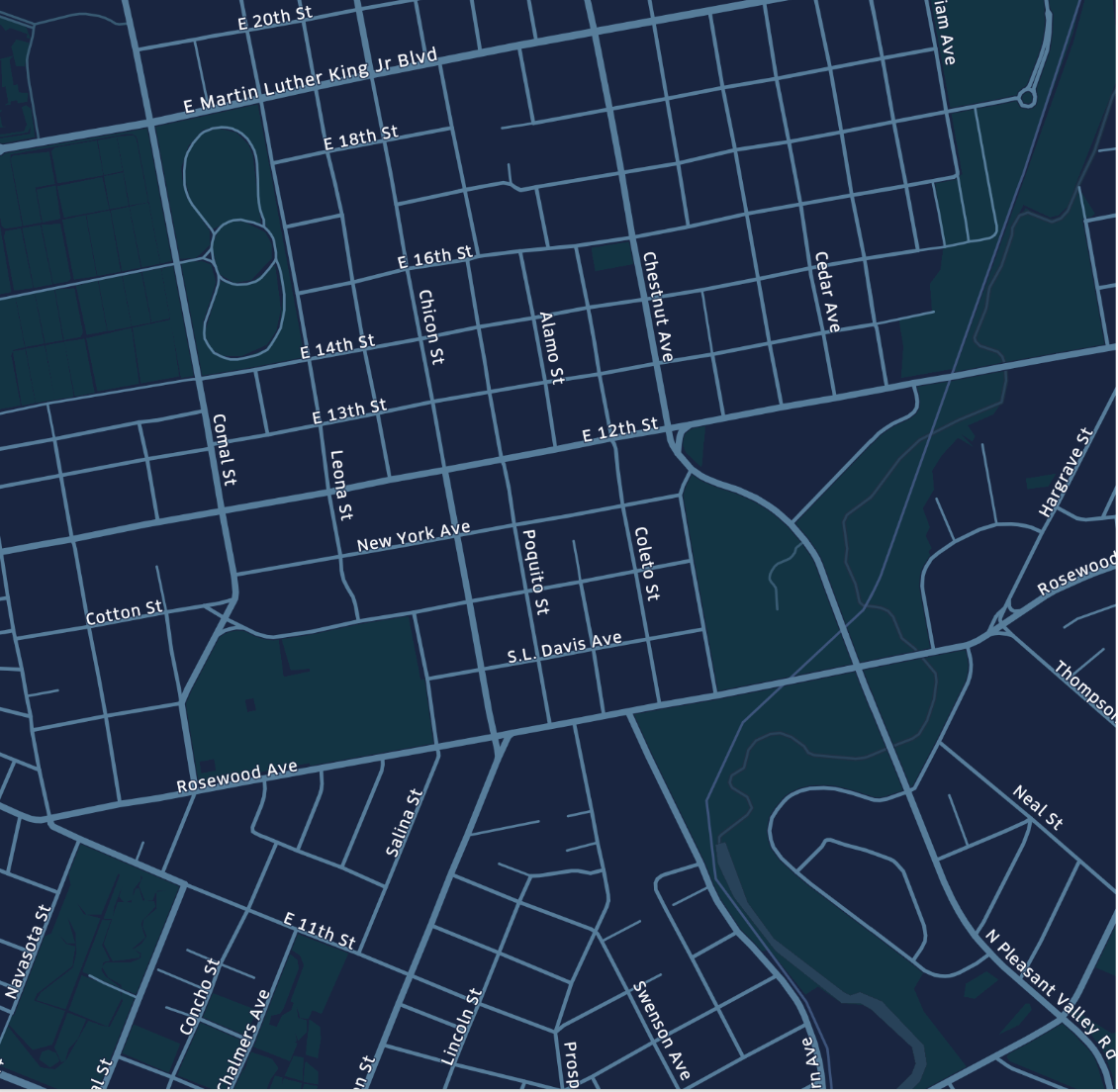}
\end{minipage}\qquad
\begin{minipage}[b]{.45\linewidth}
\includegraphics[width=\linewidth]{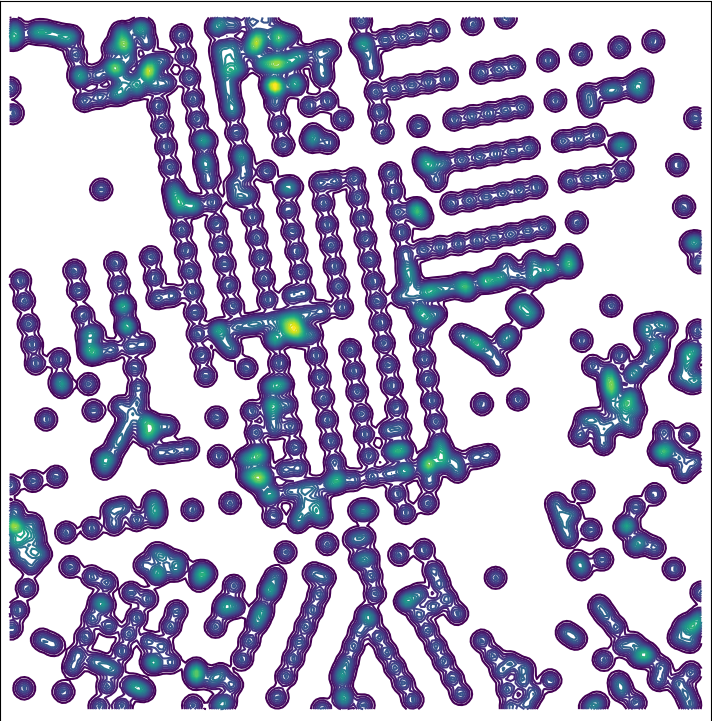}
\end{minipage}\qquad
\begin{center}
\caption{Predicted Stop sign locations (in yellow on the left and red on the right) 
for a 5x5 block region in San Francisco, CA.
Note in the figure on the right that the ground truth labels (blue crosses) for Stop signs 
are incomplete in the region.  The assignment solver has no knowledge of the ground truth clusters.}
\end{center}
\label{fig:austin}
\end{figure*}

\begin{table*}
\begin{center}
\scalebox{0.8}{
\begin{tabular}{|c|c c c c c c c c c c|}
\hline
             & & Stop & Crosswalk & DoNotEnter & NoLeft & LeftYield & NoRight & NoRightCond & NoLeftU & 2WayTraffic\\
\hline
             & Truth & 228 & 56 & 54 & 50 & 2 & 19 & 4 & 6 & 11\\
\hline
    baseline & Prec & 0.68 & 0.5 & 0.6 & 0.52 & 0.5 & {\bf 1.0} & 0.0 & 0.0 & {\bf 1.0}\\
             & Recall & {\bf 0.87} & {\bf 0.55} & {\bf 0.63} & {\bf 0.6} & {\bf 0.5} & 0.26 & 0.0 & 0.0 & 0.18\\
             & AUC & 0.74 & 0.44 & 0.55 & 0.44 & 0.57 & 0.26 & 0.0 & 0.0 & 0.18\\
\hline
    manual   & Prec & 0.82 & 0.74 & {\bf 0.88} & 0.7 & {\bf 1.0} & 0.91 & 0.2 & {\bf 0.62} & 0.67\\
             & Recall & 0.84 & 0.41 & 0.52 & 0.52 & {\bf 0.5} & {\bf 0.53} & 0.25 & {\bf 0.83} & {\bf 0.36}\\
             & AUC & 0.83 & {\bf 0.46} & \bf{0.64} & 0.46 & {\bf 0.81} & {\bf 0.55} & 0.03 & {\bf 0.77} & {\bf 0.32}\\
\hline
    SVI      & Prec & {\bf 0.84} & {\bf 0.76} & {\bf 0.88} & {\bf 0.72} & {\bf 1.0} & 0.91 & {\bf 0.5} & {\bf 0.62} & 0.67\\
             & Recall & 0.83 & 0.45 & 0.52 & 0.52 & {\bf 0.5} & {\bf 0.53} & {\bf 0.5} & {\bf 0.83} & {\bf 0.36}\\
             & AUC & {\bf 0.85} & {\bf 0.46} & 0.63 & {\bf 0.48} & {\bf 0.81} & {\bf 0.55} & {\bf 0.31} & {\bf 0.77} & {\bf 0.32}\\
\hline
\end{tabular}
}
\caption{Results on road signs for the city region in San Francisco. AUC = Area under the curve, which is
our primary metric of interest. The
baseline is the $k$-means clustering algorithm.  ``Manual" is the EM algorithm with hand tuned
parameters. ``SVI" is the EM algorithm with parameters tuned by SVI. Note that the ground truth
is incomplete so precision numbers should be higher for all three methods.}
\label{table:results}
\end{center}
\end{table*}

\begin{table*}[!htb]
\begin{center}
\scalebox{0.8}{
\begin{tabular}{|c|c c c c c c c c|}
\hline
             & & Stop & Crosswalk & Yield & StateRoute & TempParking & Merge & DoNotEnter\\
\hline
    Without priors & Prec & 0.4 & 0.60 & 0.55 & 0.12 & 0.14 & 0.43 & 0.63\\
             & Recall & {\bf 0.88} & {\bf 0.52} & {\bf 0.79} & {\bf 1.0} & {\bf 0.53} & {\bf 1.0} & {\bf 0.85}\\
\hline
    With priors & Prec & {\bf 0.42} & {\bf 0.63} & {\bf 0.61} & {\bf 0.14} & {\bf 0.17} & {\bf 0.50} & {\bf 0.71}\\
             & Recall & {\bf 0.88} & {\bf 0.52} & {\bf 0.79} & {\bf 1.0} & {\bf 0.53} & {\bf 1.0} & {\bf 0.85}\\
\hline
\end{tabular}
}
    \caption{Results on road signs with and without a road network in various regions in Austin, TX.
             Regions are all approximately 4x4 blocks and were selected based on regions where there 
             was the highest density of ground truth.}
\label{table:austin}
\end{center}
\end{table*}

We evaluate our algorithm on the precision/recall curve evaluated over an entire city, the
largest region.  During training, the algorithm only trains in the ground truth regions and for
Stop signs, only in one 5x5 region, which means that the other 9 5x5 blocks are not viewed
during training.  At test time, the models predict on all the regions together.  The results are 
shown in table ~\ref{table:results}.

There are a few important trends to note here.  The first is that our algorithm outperforms
the baseline in most circumstances with either parameter setting.  In instances where the
baseline performs better, especially in recall, the difference is often by an order of magnitude
less than the difference in which our algorithm outperforms the baseline.  The next trend to note
is that SVI tends to produce equal or better results than the initial hand tuned parameters.  This gives us
reasonable confidence that for sign types that {\it have enough ground truth for a trainable signal}, the 
algorithm will learn parameters that are consistent with real world observations.  This is further
reaffirmed by the fact that for sign types that have very few ground truth and detections such as Left Yield
or No Left U-turn, SVI learns parameters that do not differ significantly from the initialization.

The metric of particular interest is the AUC, which measures the precision-recall trade-off of these algorithms
make.  For all but one sign type, SVI has the best AUC; in one exception it is 0.01 below
the hand-tuned model.  While the baseline has relatively high recall, especially with sign types that
are abundant in the region, it has lower precision and AUC than our Bayesian algorithm.

In summary, we find that SVI successfully learns global parameter values even in the presence
of unknown data associations and limited ground truth data.
More importantly, it allows a single hierarchical generative model to generalize to different
sign types by learning parameters for each sign type.
As long as we are reasonably confident that
our generative model is faithful to real world observations, we can reuse the same model to train
a variety of clustering solvers for different traffic objects.

\subsection{Confidence Tuning}
We learn the confidence scores of the rays as a sign-specific parameter.  The upstream neural
detector often gives an unreliable confidence score, and so we learn confidence parameters
correlating the detector issued confidence and its probability of associating
with a cluster.  In figure \ref{fig:cdf}, observe that most of the rays for Stop signs
are pretty reliable; the algorithm was able to assign over 5000 rays to clusters, though
this is not the case for say Crosswalk signs.

\subsection{False detections}
False detections take the form of objects similarly colored or shaped as real
signs.  These objects exist as detections but are fewer in number than true 
detections since an object that appears to be a sign in one frame could be 
correctly classified as not a sign in later frames as the camera approaches the
object. This naturally affects rarer signs more than common signs since there
are not enough rays for the algorithm to be confident it is a false detection.
For signs such as Stop signs, the ratio true detections to false detections is higher
than that of Crosswalk signs, which have much fewer detections.

Without eccentricity and variance pruning, the curve is fairly consistent 
since all the remaining clusters are of high confidence since presumably more
detections contributed to that cluster.  Without the pruning, recall dramatically
increases at the cost of precision since clusters in the middle streets arise from
two observations that see sign behind the other.  Without thresholding, a sparse
number of detections would lead to the presence of low probability clusters.

\subsection{Intersection parameter training}\label{sec:inter}

%

We incorporate further prior information in cities where we have access
to the road network as an ablation study.
We train a MAP estimate of the intersection affinity for each sign 
type.  This training can be done completely offline (\ie outside of the SVI training
loop), as it is fitting 
parameters directly to a road network, and as such, does not require the EM solver
or the generative model used in SVI.

We use a Spike-and-slab prior over the entire area which places a Gaussian distribution
at intersections and Uniform distribution everywhere else as in figure ~\ref{fig:austin}. 
We train on a road network in Austin, TX,
and compare clustering prediction results between those with a prior over the road network 
and those without (\ie a Uniform prior across the entire region).  We learn the affinity of signs
to intersections given an intersection radius, which is the scale of the Gaussian.  These
parameters can theoretically also be transferred across cities since traffic laws are almost
identical across states which means signs are used in the same capacity 
(\eg Stop signs are located at intersections and require a complete stop for 
all cities in the US).

We test in different regions per sign type since the ground truth is sparse across the
city, with each region roughly consisting of 12-25 contiguous blocks in a rectangle.  We manually
select regions with a high concentration of ground truth and run both models identically
with the exception of the additional prior.  As displayed Table ~\ref{table:austin}, the streetmap
prior seems to not affect the recall, as both models perform identically for all sign types.
It seems to help on precision, giving strictly better results than the version without.
One explanation for this is that we use a sparse prior which is most effective in culling out
outlier clusters. A true cluster usually has many detections associated with
it, which makes it likely for the clustering algorithm to predict a sign there, regardless
of the streetmap prior \ie the prior provides weak information.  A false detection however,
often comes from a few sporadic rays that
don't necessarily converge at a single point.  In the presence of the streetmap prior we use, it can possibly
be dropped, depending on how close the candidate location is to an intersection and the sign's
affinity to be near intersections.

\section{Conclusion}
We present a framework for sensor fusion through stochastic optimization.  We introduce
an end-to-end trainable EM clustering algorithm that solves the JPDA problem, which is 
trained with a generative model through variational inference.  We demonstrate an
improvement in results through a combination of 
heuristics and additional prior information.  This technique can be used for problems
that employ bundle adjustment, and has real world impact in the development of mapping
technology.

We would like to eventually train the upstream DNN detector, in an active learning setup,
where the predicted clusters from our clustering algorithm could be used as ground truth
for the classifier. In this setup, both the detector and the clustering algorithm can be
trained in a completely unsupervised manner.

\section{Acknowledgments}
We would like to thank Martin Jankowiak, Peter Dayan, Karl Obermeyer for helpful discussions and advice,
Felipe Such for help running experiments, and the Pyro team for software support.

{\small
\bibliographystyle{ieee}
\bibliography{egbib}
}

\end{document}